\documentclass[conference]{IEEEtran}
\IEEEoverridecommandlockouts

\IEEEpubid{\makebox[\columnwidth]{Preprint; Accepted to iSAI-NLP 2025.}
\hspace{\columnsep}\makebox[\columnwidth]{ }}

\usepackage{cite}
\usepackage{amsmath,amssymb,amsfonts}
\usepackage{algorithmic}
\usepackage{graphicx}
\usepackage{textcomp}
\usepackage{multirow}
\usepackage{url}
\usepackage{xcolor}
\usepackage{booktabs}
\usepackage[font=normalsize,labelfont=bf]{caption}

\usepackage{fontspec}

\newfontfamily\padauktext[Script=Myanmar]{Padauk-Regular.ttf}
\def\BibTeX{{\rm B\kern-.05em{\sc i\kern-.025em b}\kern-.08em
    T\kern-.1667em\lower.7ex\hbox{E}\kern-.125emX}}

\def\BibTeX{{\rm B\kern-.05em{\sc i\kern-.025em b}\kern-.08em
    T\kern-.1667em\lower.7ex\hbox{E}\kern-.125emX}}

\makeatletter
\newcommand{\linebreakand}{%
  \end{@IEEEauthorhalign}
  \hfill\mbox{}\par
  \mbox{}\hfill\begin{@IEEEauthorhalign}
}
\makeatother

\begin{document}

\title{Enhancing Burmese News Classification with Kolmogorov-Arnold Network Head Fine-tuning}


\author{
Thura Aung$^{1,2}$,
Eaint Kay Khaing Kyw$^{1,2}$,
Ye Kyaw Thu$^{1,3,\ast}$,
Thazin Myint Oo$^{1}$,
Thepchai Supnithi$^{3,\ast}$
\\
$^{1}$Language Understanding Laboratory, Myanmar
\\
$^{2}$Department of Computer Engineering, KMITL, Bangkok, Thailand  
\\
$^{3}$Language and Semantic Technology Research Team, NECTEC, Bangkok, Thailand
\\
\emph{Corresponding authors:} \texttt{yekyaw.thu@nectec.or.th}, \texttt{thepchai.supnithi@nectec.or.th}
\\
\emph{Emails:} \texttt{66011606@kmitl.ac.th}, \texttt{66011533@kmitl.ac.th}, \texttt{queenofthazin@gmail.com}
}

\maketitle

\begin{abstract}
In low-resource languages like Burmese, classification tasks often fine-tune only the final classification layer, keeping pre-trained encoder weights frozen. While Multi-Layer Perceptrons (MLPs) are commonly used, their fixed non-linearity can limit expressiveness and increase computational cost. This work explores Kolmogorov–Arnold Networks (KANs) as alternative classification heads, evaluating Fourier-based FourierKAN, Spline-based EfficientKAN, and Grid-based FasterKAN—across diverse embeddings including TF-IDF, fastText, and multilingual transformers (mBERT, Distil-mBERT). Experimental results show that KAN-based heads are competitive with or superior to MLPs. EfficientKAN with fastText achieved the highest F1-score (0.928), while FasterKAN offered the best trade-off between speed and accuracy. On transformer embeddings, EfficientKAN matched or slightly outperformed MLPs with mBERT (0.917 F1). These findings highlight KANs as expressive, efficient alternatives to MLPs for low-resource language classification.
\end{abstract}


\section{Introduction}
News sentence classification involves assigning predefined categories to news text by analyzing its content and identifying meaningful patterns. Automating this task enables faster and more accurate handling of large-scale news data. In this work, we investigate head fine-tuning with different classification heads. Using the standard multilayer perceptron (MLP) head as a baseline, we evaluate three variants of the Kolmogorov–Arnold Network (KAN) \cite{liu2024kan}—FourierKAN \cite{fourierkan2025}, EfficientKAN \cite{efficientkan} and FasterKAN \cite{fasterkan}. These are tested across both static embeddings, including TF-IDF, random, and fastText \cite{bojanowski2017enriching}, and contextual embeddings derived from transformer backbones - multilingual BERT (mBERT) \cite{DBLP:journals/corr/abs-1810-04805} and Distil-mBERT \cite{Sanh2019DistilBERTAD}.

We contributed Burmese News Classification Dataset and presented the performance comparison between MLP and KAN classification heads. Our results show that EfficientKAN combined with fastText embeddings achieves the best overall performance (F1 of 0.928), while FasterKAN provides the fastest training with competitive accuracy. We further analyze trade-offs between accuracy, parameter size, and training efficiency, demonstrating that KAN-based heads are lightweight yet expressive alternatives to conventional MLPs in low-resource settings.

\begin{table*}[ht]
\centering
\renewcommand{\arraystretch}{1.4}
\caption{Example Sentences from Each News Type from News Classification Dataset.}
\begin{tabular}{|c|p{15cm}|}
\hline
\textbf{Class} & \textbf{Example Sentences} \\
\hline
Sports & \padauktext{အကြို ဗိုလ်လုပွဲ တွင် အနိုင် ရ ရှိ သည့် နှစ် သင်း က ဗိုလ်လုပွဲ ထပ်မံ ယှဉ်ပြိုင် ရ မည် ဖြစ် ပြီး ရှုံးနိမ့် သည့် နှစ် သင်း က တတိယ နေ ရာ လု ပွဲ ဆက်လက် ကစား ရ မည် ဖြစ် သည် ။} \newline The two teams that win the semi-finals will compete in the final, while the two losing teams will play in the third-place match. \\
\hline
Politics & \padauktext{အရေးအကြီးဆုံး က သူ့ ကို ခြစား မှု ၊ အလွဲသုံးစား မှု နဲ့ အမြတ်ထုတ် မှု စွပ်စွဲ မှု တွေ မ ရှိ ဖူး သေး ပါ ။ မစ္စတာ အူမဲရောဗ် ဟာ ဟော့လော့စ် ပါတီ ကနေ ရွေးကောက်ပွဲ ဝင် ရင်း နိုင်ငံ ရေး လောက ထဲ ကို ၂၀၁၉ မှာ ဝင်ရောက် လာ ခဲ့ တယ် ။ အဲဒီ နောက် အစိုးရ အရာရှိ ဖြစ် လာ ခဲ့ ပါ တယ် ။} \newline Most importantly, he has never been accused of corruption, misuse of funds, or exploitation. Mr. Umerov entered politics in 2019, running as a candidate for the Holos party. After that, he became a government official. \\
\hline
Technology & \padauktext{နိုင်ငံတကာ အာကာသ စခန်း ဟာ လာ မယ့် ၂၀၃၀ အထိ အလုပ် လုပ် နေ ဦး မှာ ဖြစ် ပြီး ၂၀၃၁ မှာ ပစိဖိတ် သမုဒ္ဒရာ ထဲ ကို ပျက်ကျ လာ လိမ့် မယ် လို့ နာဆာ က ပြော ပါ တယ် ။} \newline NASA says the International Space Station will continue to operate until 2030 and will de-orbit into the Pacific Ocean in 2031. \\
\hline
Business & \padauktext{၂၀၂၃ ၂၀၂၄ ဘဏ္ဍာ နှစ် ငါး လ အတွင်း ပြည်ပ သို့ ရေ ထွက် ပစ္စည်း တင် ပို့ မှု မှ ဒေါ်လာ ၂၄၁ သန်း ကျော် ရ ရှိ ပြီး ပင်လယ် ရေ ကြောင်း ကုန်သွယ် မှု မှ ဒေါ်လာ ၁၃၄ သန်း ကျော် တင် ပို့ ထား ကြောင်း စီးပွား ရေး နှင့် ကူးသန်းရောင်းဝယ် ရေး ဝန်ကြီးဌာန မှ သိ ရ သည် ။}  \newline The Ministry of Commerce announced that over \$241 million was earned from exporting fishery products abroad during the first five months of the 2023-2024 fiscal year, with more than \$134 million of that coming from maritime trade. \\
\hline 
Entertainment & \padauktext{အဖွဲ့ ဝင် လီဆာ အင်စတာဂရမ် တွင် ပို့စ် တစ် ခု တင် လိုက် သည် နှင့် အမေရိကန် ဒေါ်လာ ၅၇၅၀၀၀ ရ နေ ပြီး သား ဟု ဆိုရှယ်မီဒီယာ မားကတ်တင်း ပလက်ဖောင်း က ထုတ်ပြန် သည့် အင်စတာဂရမ် ဖြင့် ချမ်းသာ နေ သူ များ စာရင်း တွင် ဆို ထား သည် ။} \newline According to a list of "Instagram rich" compiled by a social media marketing platform, group member Lisa earns \$575,000 for each post on Instagram.\\
\hline
Environment & \padauktext{ဒီ မျောက် တွေ ကို အမဲ လိုက် သတ်ဖြတ် နေ တာ နဲ့ ပက်သက် ပြီး ဥပဒေ အရ အရေးယူ တာ တွေ ၊ မျောက် လေး တွေ ကို ထိန်းသိမ်း တာ တွေ ကို တော့ မ ကြား မိ သေး ဘူး လို့ ကိုဝင်းပိုင်ဦး က ဆက် ပြော ပါ တယ် ။} \newline Ko Win Paing Oo added that he hasn't heard of any legal action being taken against those who are hunting and killing the monkeys or any efforts to conserve them.\\
\hline
\end{tabular}
\label{tab:news_examples}
\end{table*}

\section{Related Work}
Previous studies on Burmese news classification have applied traditional machine learning models \cite{b1,b2} as well as deep learning approaches, including CNN \cite{b3}, BiLSTM \cite{b4}, and hybrid models \cite{b5,b6}, using both syllable- and word-level tokenization \cite{b7}. 

A widely used approach for adapting pre-trained models to downstream tasks is classification head fine-tuning, often referred to as linear probing. In this setup, only the final classification layer is updated while the underlying backbone remains fixed. This strategy is particularly effective in low-resource scenarios \cite{gao2023tuning} and has been shown to maintain stronger robustness under out-of-distribution shifts \cite{kumar2022fine} compared with full-model fine-tuning.

In natural language processing tasks such as text classification, it is common to place a Multi-Layer Perceptron (MLP) head \cite{hornik1989multilayer} on top of transformer-based backbones \cite{vaswani2017attention}. Despite their popularity, MLP heads may fail to fully exploit the contextual richness of pre-trained embeddings and often introduce significant computational overhead. Recent studies \cite{imran2024fourierkan} have proposed FourierKAN, a KAN varient, which leverage Fourier-based transformations and have demonstrated superior performance over conventional MLP heads in fine-tuning tasks. 

\begin{table}[h]
    \renewcommand{\arraystretch}{1}
    \centering
    \small
    \caption{Label Counts and Percentages}
    \label{tab:label_counts}
    \begin{tabular}{|c|r|r|}
    \hline
    \textbf{Class} & \textbf{Count} & \textbf{Percentage} \\ 
    \hline
    Sports & 1,232 & 16.84\% \\ 
    Politics & 1,228 & 16.79\% \\ 
    Technology & 1,224 & 16.73\% \\ 
    Business & 1,221 & 16.69\% \\ 
    Entertainment & 1,205 & 16.47\% \\ 
    Environment & 1,205 & 16.47\% \\ 
    \hline
    \end{tabular}
\end{table}

\section{Dataset Preparation}
We collected Burmese news paragraphs across six categories from various news websites and manually labeled them, as summarized in Table \ref{tab:label_counts}. The news articles were gathered from Burmese news websites such as VOA Burmese\footnote{\url{https://burmese.voanews.com/}}, BBC Burmese\footnote{\url{https://www.bbc.com/burmese}}, and Radio Free Asia (RFA)\footnote{\url{https://www.rfa.org/burmese/}}, and annotated by native speakers between April and June 2024.  The dataset was split into training 80\% (5.84k sentences) and testing 20\% (1.47k sentences) of the original dataset.

Prior to experimentation, the text underwent cleaning and normalization based on syllable rules to ensure correct Unicode ordering. Proper ordering is crucial for accurate word and sentence segmentation \cite{aung2023mysentence, thu2023neural} and for consistent model training. After preprocessing, we performed tokenization using mBert tokenizer for transformer-based approaches. For static embedding-based models, we used the myWord\footnote{\url{https://github.com/ye-kyaw-thu/myWord}} \cite{thu2021myword} tool, which utilizes unigram and bigram dictionaries derived from 0.5 million sentences and 12 million words. Figure \ref{tab:news_examples} shows examples of each class in our news classification dataset.

\section{Methodology}
\subsection{Embeddings}

In this work, we focus on evaluating the effectiveness of different embedding strategies combined with different classification heads. Neural embeddings map discrete tokens into continuous vectors that capture semantic and syntactic relationships. We compare static embeddings and contextual embeddings while keeping the backbone frozen and fine-tuning only the classification head. 

For static embeddings, we experiment with TF-IDF, random embeddings, and fastText \footnote{\url{facebook/fasttext-my-vectors}} \cite{bojanowski2017enriching} embeddings. These embeddings provide fixed representations for each token, which are then fed into a classification head. For contextual embeddings, we use pre-trained models such as multilingual BERT (mBERT\footnote{\url{google-bert/bert-base-multilingual-cased}}) \cite{DBLP:journals/corr/abs-1810-04805}  and multilingual distilled Bert (Distil-mBERT\footnote{\url{distilbert/distilbert-base-multilingual-cased}}) \cite{Sanh2019DistilBERTAD}. 

\begin{figure}
    \centering
    \includegraphics[width=0.9\linewidth]{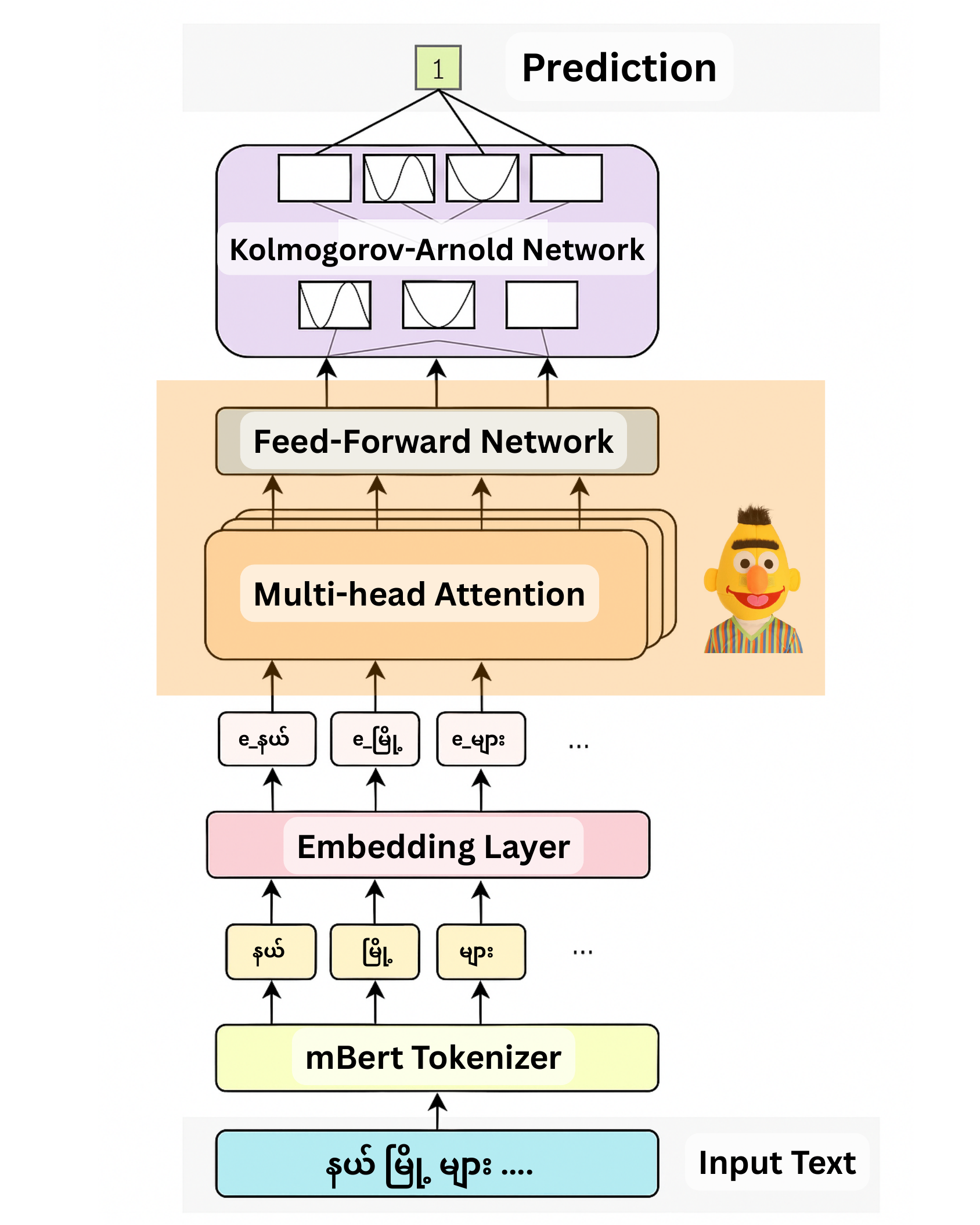}
    \caption{Integration of KAN as a classification head on top of a Transformer encoder for context-aware prediction}
    \label{fig:archi}
\end{figure}

These embeddings dynamically encode each token based on its context in the input sequence and are also passed through a classification head without further updating the backbone. As illustrated in Figure~\ref{fig:archi}, the input text is first processed by a pretrained Transformer encoder to obtain contextual embeddings for each token. These embeddings are then passed to the KAN layer, which is trained to serve as the classification head while keeping the encoder backbone fixed. The trained KAN layer maps the contextual representations to their corresponding predicted classes.

\subsection{Models}

\subsubsection{Baseline MLP}
We use a standard Multi-Layer Perceptron (MLP) as a baseline classifier.  
For a 1-layer MLP:  
\begin{equation}
\hat{y} = \text{softmax}\bigl(W_0 H + b_0\bigr)
\end{equation}
and for a 2-layer MLP:  
\begin{align}
h_0 &= \sigma(W_0 H + b_0), \\
\hat{y} &= \text{softmax}(W_1 h_0 + b_1),
\end{align}
where $\sigma(x) = 1 / (1 + e^{-x})$ is the sigmoid activation, $H$ is the input embedding, and $W_i, b_i$ are trainable parameters.

\subsubsection{Kolmogorov-Arnold Representation Theorem (KART)}
KART \cite{schmidt-hieber2021kolmogorov, selitskiy2022kolmogorov} states that any multivariate continuous function can be decomposed into sums of univariate functions:  
\begin{equation}
f(\mathbf{x}) = \sum_{q=1}^{2n+1} \Phi_q \Biggl( \sum_{p=1}^n \phi_{q,p}(x_p) \Biggr),
\end{equation}
where $\phi_{q,p}$ are simple single-variable functions applied to each input $x_p$, and $\Phi_q$ are outer functions combining these sums. This decomposition motivates KANs, which use learnable $\phi_{q,p}$ and $\Phi_q$ for adaptive modeling.

\subsubsection{Kolmogorov-Arnold Network (KAN)}
KAN layers extend KART by parameterizing $\phi_{q,p}$ and $\Phi_q$ as learnable splines. For a single KAN layer with $n_\text{in}$ input and $n_\text{out}$ output dimensions, the layer output is:  
\begin{equation}
x^{(l+1)}_j = \sum_{i=1}^{n_l} \phi^{(l)}_{j,i}\bigl(x^{(l)}_i\bigr), \quad j = 1, \dots, n_{l+1},
\end{equation}
or in matrix form:
\begin{equation}
\mathbf{x}^{(l+1)} = \Phi^{(l)} \mathbf{x}^{(l)}
\end{equation}
where $\Phi^{(l)}$ is the function matrix of learnable splines for the $l$-th layer. The approximation error is bounded by:  
\begin{equation}
\| f - (\Phi_G^{(L-1)} \circ \dots \circ \Phi_G^{(0)}) \mathbf{x} \|_{C^m} \le C G^{-k-1+m},
\end{equation}
with grid size $G$, spline order $k$, and constant $C$.

\noindent
\textbf{FourierKAN} FourierKAN \cite{imran2024fourierkan} replaces spline-based activations with Fourier series expansions to represent the univariate functions $\phi_{q,p}(x_p)$ in the Kolmogorov–Arnold decomposition. Each function is approximated as a weighted sum of sine and cosine terms:
\begin{equation}
\phi_{q,p}(x_p) = \sum_{k=1}^{G} a_{q,p,k} \cos(kx_p) + b_{q,p,k} \sin(kx_p),
\end{equation}
where $G$ is the grid size (number of Fourier modes), and $a_{q,p,k}, b_{q,p,k}$ are learnable Fourier coefficients.
This formulation captures smooth periodic and non-linear patterns efficiently while maintaining a compact parameterization. The FourierKAN layer thus computes:
\begin{equation}
x^{(l+1)}_j = \sum_{i=1}^{n_l} \bigl[ A^{(l)}_{j,i} \cos(kx^{(l)}_i) + B^{(l)}_{j,i} \sin(kx^{(l)}_i) \bigr] + \beta_j,
\end{equation}
where $\beta_j$ denotes an optional bias term. Compared to spline-based KANs, FourierKAN offers smoother gradient flow, easier initialization, and efficient modeling of oscillatory or compositional data.

\renewcommand{\arraystretch}{1.4}
\begin{table*}[h!]
\centering
\caption{Efficiency comparison of models with different embeddings}
\begin{tabular}{|l|r|r|r|r|}
\hline
\textbf{Model} & \textbf{Params} & \textbf{Train (s)} & \textbf{Fwd (ms)} & \textbf{Bwd (ms)} \\
\hline
Tf-IDF + MLP                               & 0.13M     & 10.7     & 0.26     & 0.64 \\
Tf-IDF + FourierKAN                        & 2.99M     & 28.1     & 0.56     & 1.23 \\
Tf-IDF + EfficientKAN                      & 1.29M     & 21.4     & 4.49     & 4.18 \\
Tf-IDF + FasterKAN                         & 1.03M     & 7.6      & 0.58     & 1.00 \\
\hline
Random Embedding + MLP                     & 2.00M     & 13.5     & 0.44     & 0.97 \\
Random Embedding + FourierKAN              & 0.27M     & 8.0      & 0.56     & 1.23 \\
Random Embedding + EfficientKAN            & 2.35M     & 24.5     & 1.76     & 2.82 \\
Random Embedding + FasterKAN               & 2.27M     & 14.2     & 0.78     & 1.41 \\
\hline
fastText + MLP                             & 2.00M     & 18.1     & 0.46     & 0.98 \\
fastText + FourierKAN                      & 0.63M     & 8.3      & 1.08     & 1.90 \\
fastText + EfficientKAN                    & 2.35M     & 24.6     & 1.82     & 2.90 \\
fastText + FasterKAN                       & 2.27M     & 15.6     & 0.75     & 1.31 \\
\hline
Distil-mBERT + MLP                         & 135M      & 669.5    & 100.23   & 211.76 \\
Distil-mBERT + FourierKAN                  & 137M   & 795.2    & 52.09    & 113.49 \\
Distil-mBERT + EfficientKAN                & 137M      & 671.0    & 100.13   & 212.90 \\
Distil-mBERT + FasterKAN                   & 136M      & 669.4    & 99.92    & 211.73 \\
\hline
mBERT + MLP                                & 178M      & 1284.8   & 203.03   & 418.72 \\
mBERT + FourierKAN                         & 180M   & 1481.2   & 115.33   & 215.38 \\
mBERT + EfficientKAN                       & 180M      & 1291.2   & 203.57   & 420.01 \\
mBERT + FasterKAN                          & 179M      & 1289.8   & 202.75   & 418.26 \\
\hline
\end{tabular}
\label{table:efficiency}
\end{table*}

\noindent
\textbf{Efficient KAN} In spline-based KAN, sparsity, an important property for interpretability, was originally enforced via L1 regularization on the expanded input tensors \cite{liu2024kan}. In EfficientKAN, this is replaced with L1 regularization on the weights, maintaining computational efficiency while encouraging sparsity. The learnable scaling for each activation function can be optionally enabled or disabled, providing a trade-off between accuracy and efficiency.

\noindent
\textbf{FasterKAN} The original KAN expands the input tensor to shape $(\text{batch\_size}, \text{out\_features}, \text{in\_features})$ in order to apply learnable activations, which incurs high memory usage. EfficientKAN \cite{efficientkan} addresses this by first applying the activation functions to the input and then performing a linear combination of the results. This reformulation significantly reduces memory consumption and allows forward and backward computations to be implemented as standard matrix multiplications. For the activation basis, the FasterKAN authors \cite{fasterkan} investigate Reflectional Switch Activation Functions (RSWAF):  
\begin{equation}
b_i(u) = 1 - \tanh^2\Big(\frac{u - u_i}{h}\Big),
\end{equation}
which approximate spline functions, are computationally inexpensive, and can be evaluated on uniform grids. In addition, we explore alternative basis functions, different exponent and $h$ values, and the exclusion of SiLU in our experiments.

Finally, careful initialization of the weight parameters using Kaiming uniform initialization has been found to improve convergence and overall performance. Combining these strategies results in the FasterKAN variant, which integrates both computational efficiency and competitive predictive performance.

\section{Experimental Setup}
We conducted experiments on Google Colab using an NVIDIA Tesla T4 GPU with 16 GB VRAM. The models were implemented in PyTorch, with supporting libraries including scikit-learn for evaluation metrics and Hugging Face Transformers for pre-trained multilingual backbones. 

For models based on static embeddings, we employed TF-IDF, random, and fastText representations. For models with transformer-backbones, we evaluated three multilingual architectures: mBERT, and Distil-mBERT. On top of each backbone, we integrated different classification heads to compare their effectiveness.

\begin{itemize}
    \item \textbf{MLP}: Feed-forward network with ReLU activations.
    \item \textbf{FourierKAN}: Fourier-based KAN variant with grid size of 8, using sine and cosine basis functions.
    \item \textbf{EfficientKAN}: spline-based variant with grid size of 8 and cubic spline order.
    \item \textbf{FasterKAN}: grid-based KAN variant with learnable grid and inverse denominator formulation.
\end{itemize}

The training used the AdamW optimizer with a two-tier learning rate: 2e-5 for the backbone parameters and 2e-4 for the classification head, along with a Cosine Annealing scheduler. The cross-entropy loss function was used. Training was performed for 15 epochs with batch size 32 for static embeddings approaches and 5 epochs with batch size 8 for transformer-backbone models, and gradient clipping (max-norm = 1.0) was applied to stabilize training.

Model accuracy was assessed on the held-out test set, with weighted F1-score reported as the evaluation metric. To assess computational efficiency, we measured the number of parameters (total and trainable), training time, and average forward/backward pass latency (in milliseconds). Early stopping with patience of three validation checks and dropout rate of 0.3 was applied to avoid overfitting.

\begin{figure*}
    \centering
    \includegraphics[width=\linewidth]{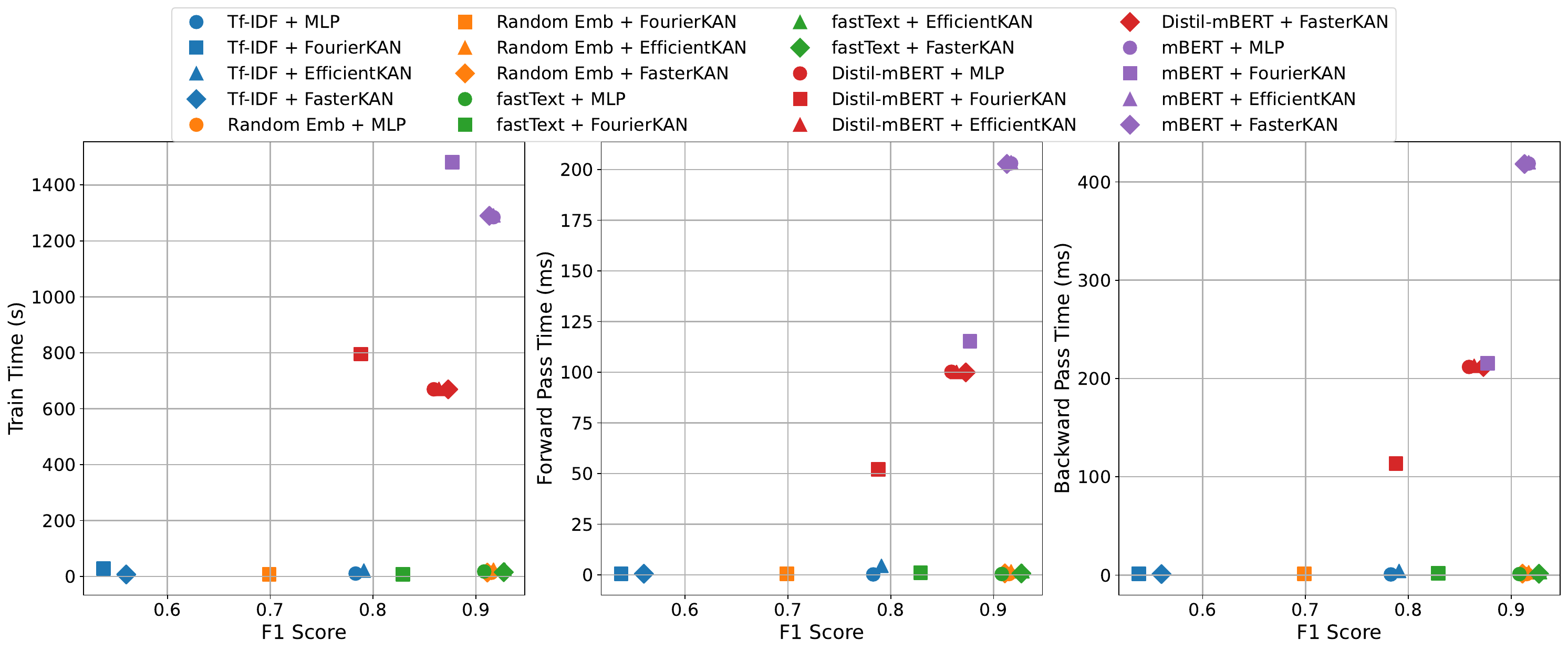}
    \caption{Comparison of model performance versus computational cost.}
    \label{fig:placeholder}
\end{figure*}

\renewcommand{\arraystretch}{1.4}
\begin{table}[h!]
\centering
\caption{F1-score comparison of classification heads across embeddings}
\begin{tabular}{|l|r|}
\hline
\textbf{Model} & \textbf{F1-score} \\
\hline
Tf-IDF + MLP                       & 0.783 \\
Tf-IDF + FourierKAN                & 0.538 \\
\textbf{Tf-IDF + EfficientKAN}     & \textbf{0.791} \\
Tf-IDF + FasterKAN                 & 0.560 \\
\hline
Random Embedding + MLP             & 0.915 \\
Random Embedding + FourierKAN      & 0.699 \\
\textbf{Random Embedding + EfficientKAN} & \textbf{0.917} \\
Random Embedding + FasterKAN       & 0.911 \\
\hline
fastText + MLP                     & 0.908 \\
fastText + FourierKAN              & 0.829 \\
\textbf{fastText + EfficientKAN}   & \textbf{0.928} \\
fastText + FasterKAN               & 0.927 \\
\hline
Distil-mBERT + MLP                 & 0.859 \\
Distil-mBERT + FourierKAN          & 0.788 \\
Distil-mBERT + EfficientKAN        & 0.864 \\
\textbf{Distil-mBERT + FasterKAN}  & \textbf{0.873} \\
\hline
\textbf{mBERT + MLP}               & \textbf{0.917} \\
mBERT + FourierKAN                 & 0.877 \\
\textbf{mBERT + EfficientKAN}      & \textbf{0.917} \\
mBERT + FasterKAN                  & 0.913 \\
\hline
\end{tabular}
\label{table:performance}
\end{table}

\section{Result and Discussion}

Figure \ref{fig:placeholder} illustrates the trade-off between model performance and computational cost across various model and embedding configurations. Each subplot plots the F1-score on the x-axis against a different computational metric on the y-axis: training time, forward pass time, and backward pass time, respectively (with lower values indicating better efficiency). The color coding distinguishes embedding types — blue for TF-IDF baseline, orange for random learned embeddings, green for fastText embeddings, and red for multilingual Transformer embeddings — while marker shapes differentiate the classification heads: circles for MLP, squares for FasterKAN, triangles for EfficientKAN, and diamonds for FourierKAN. 

\subsection{Efficiency Comparison}

\noindent
\textbf{Training Time}
Table~\ref{table:efficiency} shows that MLP serves as the baseline and is generally the fastest to train across embeddings. Interestingly, FasterKAN often trains quicker than MLP despite having more parameters, indicating an efficient architecture. In contrast, EfficientKAN consistently requires significantly longer training times regardless of the embedding type. Transformer-based models, including Distil-mBERT and mBERT, are the slowest, with training durations ranging from several hundred seconds to over one thousand seconds.



\noindent
\textbf{Inference and Backward Propagation Efficiency}
Across all models, MLP consistently demonstrates the lowest latency during both forward inference (0.26–0.46 ms) and backward propagation (0.64–0.98 ms), making it the most computationally efficient choice. FourierKAN models generally have slightly higher times than MLP but remain efficient, with forward passes around 0.56–1.08 ms and backward passes between 1.23–1.90 ms depending on the embedding. FasterKAN introduces moderate overheads, with forward times of 0.58–0.78 ms and backward times of 1.0–1.41 ms, while EfficientKAN is the slowest among non-transformer models, ranging from 1.76 to 4.49 ms forward and up to 4.18 ms backward. Transformer-based models incur substantially higher costs, with Distil-mBERT forward and backward passes around 52–100 ms and 113–212 ms respectively, and mBERT exceeding 115 ms forward and 215–418 ms backward, reflecting their greater computational demands during inference.

\subsection{Performance Comparison}

\noindent
\textbf{Effect of Classification Head}
Table~\ref{table:performance} shows EfficientKAN generally achieves the highest F1-scores for classical embeddings, followed by MLP and FasterKAN, with FourierKAN trailing behind. For TF-IDF, EfficientKAN scores 0.791 versus FourierKAN’s 0.538. Similar trends hold for Random and fastText embeddings. For transformers, FasterKAN or MLP tend to perform best, with FourierKAN slightly lower but competitive.

\noindent
\textbf{Effect of Embedding Type}
Embedding choice strongly affects performance. Learned embeddings (Random, fastText, Fourier) consistently outperform TF-IDF across all heads. fastText usually yields the highest F1, followed by Random and Fourier. Transformer embeddings provide additional gains, but improvements are smaller compared to switching from TF-IDF to learned embeddings.

Overall, FourierKAN generally underperforms compared to spline-based EfficientKAN and grid-based FasterKAN in terms of F1-score, due to architectural limitations. Its use of fixed Fourier basis functions reduces its ability to model complex, localized patterns, whereas spline-based KANs learn adaptive piecewise functions that better approximate nonlinear decision boundaries. While FourierKAN often has more trainable parameters and longer training times than FasterKAN, it is typically faster in inference due to the simplicity of its basis functions. In contrast, FasterKAN uses a grid-based architecture with RSWAF to efficiently approximate B-splines, offering a strong balance between accuracy and efficiency. Overall, despite its speed in inference, FourierKAN’s lower accuracy and higher training cost make it less favorable for most classification tasks.

\section{Conclusion and Future Work}
This work explored Kolmogorov-Arnold Networks (EfficientKAN, FasterKAN, and FourierKAN) as classification heads for Burmese news sentence classification using static and contextual embeddings.

Our results show that embedding type has the strongest impact on performance. Learned embeddings like random and fastText outperform TF-IDF, while mBERT and Distil-mBERT achieve the highest overall accuracy. Among classification heads, EfficientKAN consistently performs best with static embeddings, especially fastText (0.928 F1), while FasterKAN offers a strong balance of speed and accuracy. FourierKAN, though efficient, underperforms due to its less adaptive basis and larger parameter counts. For transformer embeddings, differences between heads are smaller, with MLP, EfficientKAN, and FasterKAN performing similarly. MLP remains fastest overall, while EfficientKAN is the most computationally expensive among non-transformer heads. 

KAN-based heads show promise, particularly with static embeddings. Future work will extend these models to other tasks in low-resource settings, explore new basis functions, and release code and data for reproducibility.

\bibliographystyle{IEEEtran}
\bibliography{references}

\end{document}